\documentclass[9pt,twocolumn,twoside]{pnas-new}

\templatetype{pnasresearcharticle} 

\begin{document}

\title{Do LLMs write like humans? Variation in grammatical and rhetorical styles}

\author[a,1]{Alex Reinhart}
\author[b]{Ben Markey}
\author[c]{Michael Laudenbach}
\author[a,d]{Kachatad Pantusen}
\author[a]{Ronald Yurko}
\author[a]{Gordon Weinberg}
\author[b]{David West Brown}

\affil[a]{Department of Statistics \& Data Science, Carnegie Mellon University, Pittsburgh, PA 15213}
\affil[b]{Department of English, Carnegie Mellon University, Pittsburgh, PA 15213}
\affil[c]{Department of Humanities \& Social Sciences, New Jersey Institute of Technology, Newark, NJ 07102}
\affil[d]{Heinz College of Information Systems and Public Policy, Carnegie Mellon University, Pittsburgh, PA 15213}

\leadauthor{Reinhart}

\significancestatement{As large language models (LLMs) have grown in power and become more widely available, research has focused on their ability to complete various tasks and the biases they exhibit when doing so. In this study, we instead examine their writing style in detail. We show that instruction-tuned models, which are trained to answer questions and solve problems, have a distinct noun-heavy, informationally dense writing style, even when prompted to match the style of informal speech and writing. These findings suggest that instruction-tuned models generate text that does not align with genre conventions familiar to human audiences, and demonstrate the value of linguistic variables in evaluating the output of LLMs.} 




\authorcontributions{A.R., D.W.B, B.M., M.L., R.Y., and G.W. designed research; A.R., D.W.B., K.P., and R.Y. performed research; A.R., D.W.B., B.M., M.L., R.Y., and G.W. wrote the paper.}
\authordeclaration{The authors declare no competing interests.}
\correspondingauthor{\textsuperscript{1}To whom correspondence should be addressed. E-mail: areinhar@stat.cmu.edu}


\begin{abstract}
Large language models (LLMs) are capable of writing grammatical text that follows instructions, answers questions, and solves problems. As they have advanced, it has become difficult to distinguish their output from human-written text. While past research has found some differences in features such as word choice and punctuation, and developed classifiers to detect LLM output, none has studied the rhetorical styles of LLMs.
Using several variants of Llama 3 and GPT-4o, we construct two parallel corpora of human- and LLM-written texts from common prompts. Using Douglas Biber's set of lexical, grammatical, and rhetorical features, we identify systematic differences between LLMs and humans and between different LLMs. These differences persist when moving from smaller models to larger ones, and are larger for instruction-tuned models than base models. This observation of differences demonstrates that despite their advanced abilities, LLMs struggle to match human stylistic variation. Attention to more advanced linguistic features can hence detect patterns in their behavior not previously recognized.

\end{abstract}

\dates{This manuscript was compiled on \today}
\doi{\url{www.pnas.org/cgi/doi/10.1073/pnas.XXXXXXXXXX}}

\maketitle
\thispagestyle{firststyle}
\ifthenelse{\boolean{shortarticle}}{\ifthenelse{\boolean{singlecolumn}}{\abscontentformatted}{\abscontent}}{}

\firstpage[3]{4}


\dropcap{A}s large language models (LLMs) have advanced in recent years, from ``stochastic parrots'' to models evidently capable of performing complex tasks, most attention has focused on their reasoning performance: solving mathematical problems, writing code, evaluating arguments, diagnosing diseases, and so on \cite{Wang:2023,Huang:2023,Chen:2023,Lehr:2024}. While past research has studied their mastery of basic grammar and vocabulary \cite{Chang:2024nl}, there is relatively little research on their language performance more generally: their ability to produce readable text in a variety of styles. Rather than exploring it in detail, commentators discuss the business and communication tasks that might be automated by LLMs with their writing ability, or consider the dangers of impersonation and misinformation facilitated by LLMs \cite{Barman:2024,Kumar:2023,Kovacs:2024yr,Hagendorff:2024}. As more and more writing tasks are automated, such problems appear inevitable.

However, the impression that LLMs write ``like humans'' is based primarily on qualitative evaluation of their output, not on thorough linguistic evaluation of their text. So far, quantitative comparisons have looked mainly at basic grammar and syntax \cite{Chang:2024nl} or features such as word choice, punctuation, sentence length, and so on, finding evidence of some differences between human- and LLM-written text \cite{Tang_2024,Frohling_2021,Munoz-Ortiz:2023kf,Herbold:2023iz,Liang:2024}. Other work has used these features, or language models trained on sample texts, to classify LLM-written texts with varying degrees of success \cite{Liu:2024os,Mosca:2023,Wang:2024}. Though not definitive, these results suggest there are indeed structural differences between human- and LLM-written text.

We used several recent LLMs (OpenAI's GPT-4o and GPT-4o Mini, and four variants of Meta Llama 3) to generate text from prompts drawn from a large, representative corpus of English, allowing us to directly compare the style of LLM writing to human writing. We find large differences in grammatical, lexical, and stylistic features, demonstrating that LLMs prefer specific grammatical structures and struggle to match the stylistic variation present in human communication, particularly as that variation aligns with the conventions that structure genres such as academic writing, interactive speech, or journalistic news. In Llama 3, where we are able to compare base models (which produce text completions) to instruction-tuned variants (which have been further trained to answer questions and complete tasks specified in prompts), we further see that the instruction tuning introduces more extreme grammatical differences, making them easier to distinguish from human writing and introducing features similar to those present in GPT-4o and GPT-4o Mini.

For example, the instruction-tuned LLMs used present participial clauses at 2 to 5 times the rate of human text, such as in this sentence from GPT-4o using two present participles: ``Bryan, \textit{leaning on his agility}, dances around the ring, \textit{evading Show's heavy blows}.'' They also use nominalizations at 1.5 to 2 times the rate of humans, such as in this sentence from Llama 3 70B Instruct containing four: ``These schemes can help to reduce \textit{deforestation}, habitat \textit{destruction}, and \textit{pollution}, while also promoting sustainable \textit{consumption} patterns.'' On the other hand, GPT-4o uses the agentless passive voice at roughly half the rate as human texts---but in each case, the Llama base models use these features at rates more closely matching humans. This suggests that instruction tuning, rather than training the models to write even more like humans, instead trains them in a particular informationally dense, noun-heavy style, and limits their ability to mimic other writing styles leading to, in some cases, genre misalignment.

These results demonstrate the value of attending to linguistic structure (morphosyntactic, functional, and rhetorical) in order to better understand the affordances and outputs of large language models. Since the rise of the Internet and the concurrent development of efficient processing architectures, language modeling has relied on relatively simple linguistic principles (i.e., sequences and context windows) as the availability of massive amounts of text allowed models trained on text to rapidly outperform older paradigms based on linguistic theory \cite{Li:2023}; but linguistic theory can provide better ways to evaluate LLM output, just as improved benchmark problems can provide better ways to evaluate their reasoning ability. These results also demonstrate the limits of current LLMs in matching human language, showing that despite their apparent ability, they have measurable limitations compared to human authors.

\section*{Methods}

\begin{figure}
    \centering
    \begin{tikzpicture}[>=latex]
        \node(chunk1) at (0, 2) {Chunk 1};
        
        \node(prompt) at (0, 0) {LLM prompt};

        \begin{scope}[text width=3.5cm, align = left]
            \node(chunk2) at (5, 2) {Chunk 2};
            \node(llama8b) at (5, 1) {Llama 3 8B};
            \node(llama8binstruct) at (5, 0.5) {Llama 3 8B Instruct};

            \node(llama70b) at (5, 0) {Llama 3 70B};
            \node(llama70binstruct) at (5, -0.5) {Llama 3 70B Instruct};

            \node(gpt4o) at (5, -1) {GPT-4o};
            \node(gpt4omini) at (5, -1.5) {GPT-4o Mini};
        \end{scope}

        \draw[->] (chunk1) -- (chunk2);
        \draw[->] (chunk1) -- (prompt);

        \draw[->] (prompt) -- (llama8b.west);
        \draw[->] (prompt) -- (llama8binstruct.west);
        \draw[->] (prompt) -- (llama70b.west);
        \draw[->] (prompt) -- (llama70binstruct.west);
        \draw[->] (prompt) -- (gpt4o.west);
        \draw[->] (prompt) -- (gpt4omini.west);
    \end{tikzpicture}
    \caption{The LLM text generation workflow. Each human text was split into two chunks of roughly 500 words; the first chunk was used to prompt an LLM to create text that was compared to the second human chunk.}
    \label{fig:workflow}
\end{figure}
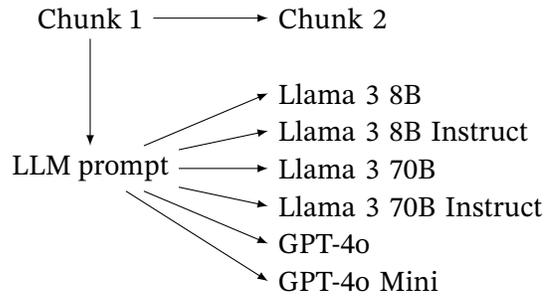

We created two corpora of parallel human- and LLM-written texts. Each corpus began with $n = 12,000$ human-authored English texts from a range of genres, from spoken word (such as podcast transcripts) to news and magazine articles to formal academic writing. Language use varies in relation to situational factors such as audience and purpose \cite{Biber:1988, Miller:1984}; by including multiple genres, we aimed to capture a diverse range of language production. As illustrated in Figure~\ref{fig:workflow}, from each text we extracted two consecutive chunks of roughly 500 words (split at sentence boundaries). The first chunk was provided to each LLM to give it context and a sample of the writing style. The LLMs were prompted to write 500 more words in the same style, tone, and diction; their generated text was then compared to the next 500-word chunk of the human text. We used six LLMs: GPT-4o and GPT-4o Mini \cite{GPT4o,GPT4oMini} and Meta Llama 3 8B, 70B, 8B Instruct, and 70B Instruct \cite{Llama3}, producing six LLM-authored texts for each human-authored text. See the SI Appendix for detailed prompt information.

We constructed the first corpus, the Human-AI Parallel English corpus, from six categories of text (academic, news, fiction, spoken word, blogs, and TV/movie scripts). The second corpus, the COCA AI Parallel (CAP) Corpus, is drawn from the pre-existing Corpus of Contemporary American English (COCA), a large, representative corpus of over 1 billion words in eight registers: spoken, fiction, magazines, newspapers, academic, blogs, web pages, and TV/movie subtitles \cite{COCA}. The HAP-E corpus was used for our primary analyses, while CAP was used to evaluate the generalizability of the results to different texts. The LLMs sometimes refused to respond to prompts or gave short, unusable answers; after these were removed, there were $n = 8,290$ HAP-E texts and $n = 9,615$ CAP texts with outputs from all LLMs. With two human chunks and six LLM-authored chunks for each text, HAP-E comprised $n=66,320$ chunks and CAP $n=76,920$ chunks. See SI Appendix Tables S2--S3 for corpus size and composition.

To extract meaningful features from our corpus for training our classifiers, we used Douglas Biber’s tagset of 66 linguistic categories \citep{Biber:1988,Biber:1995,Biber:2009}, which includes indices of lexical complexity and raw linguistic features ranging from the lexical to the grammatical. For example, features include mean word length, the use of nominalizations (nouns formed from adjectives or verbs, such as \textit{development} or \textit{robustness}), agentless passive voice, hedging phrases (such as \textit{something like} or \textit{almost}), and clausal coordination. All features are listed in the SI Appendix, Table S4. Differences between LLM and human use of features were tested for statistical significance with the paired Wilcoxon signed-rank test with Bonferroni multiple comparison correction.

As a further check of generalizability, we used part of the M4 parallel corpus \cite{Wang:2024} consisting of abstracts from the arXiv preprint service alongside abstracts generated by GPT-3.5 when prompted with the preprint title. These texts are from a different LLM (GPT-3.5) and a very distinct genre of writing (academic abstracts), providing a check on the consistency of results in different genres.

\section*{Results}

\subsection*{Classifying text by source}

\begin{figure}
    \centering
    \includegraphics[width=\linewidth]{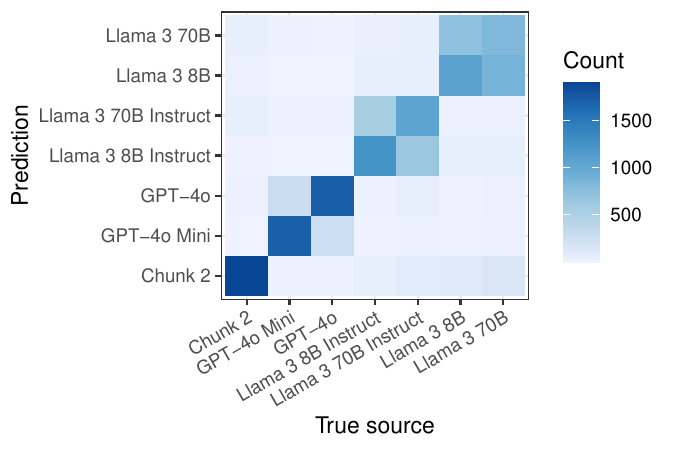}
    \caption{Confusion matrix for a random forest classifying HAP-E texts by their linguistic and rhetorical features, evaluated on the test set (25\% of the HAP-E corpus, including $n = 14,535$ human and LLM texts). The block diagonal structure indicates that most classification errors were between different versions of the same LLM, rather than between humans and LLMs.}
    \label{fig:confusion-matrix}
\end{figure}

A random forest classifier using the Biber features to distinguish between the seven text sources in HAP-E (human chunk 2 and the six LLMs) achieved a test accuracy of 66\%, compared to an expected accuracy of 14\% from random guessing. The confusion matrix, shown in Figure~\ref{fig:confusion-matrix}, demonstrates that little of the error was due to confusion between human texts and the LLMs: instead, most classification errors confused Llama 3 8B and 70B, Llama 3 8B Instruct and 70B Instruct, or GPT-4o and 4o Mini. Each pair consists of models of two different sizes trained on similar data, implying that the size difference does not produce dramatically different style. Overall, only 4.2\% of LLM texts were falsely classified as human, and only 9.8\% of human texts were falsely classified as LLMs.

\subsection*{Differences in style and vocabulary}

\begin{figure*}
    \centering
    \includegraphics[width=0.7\textwidth]{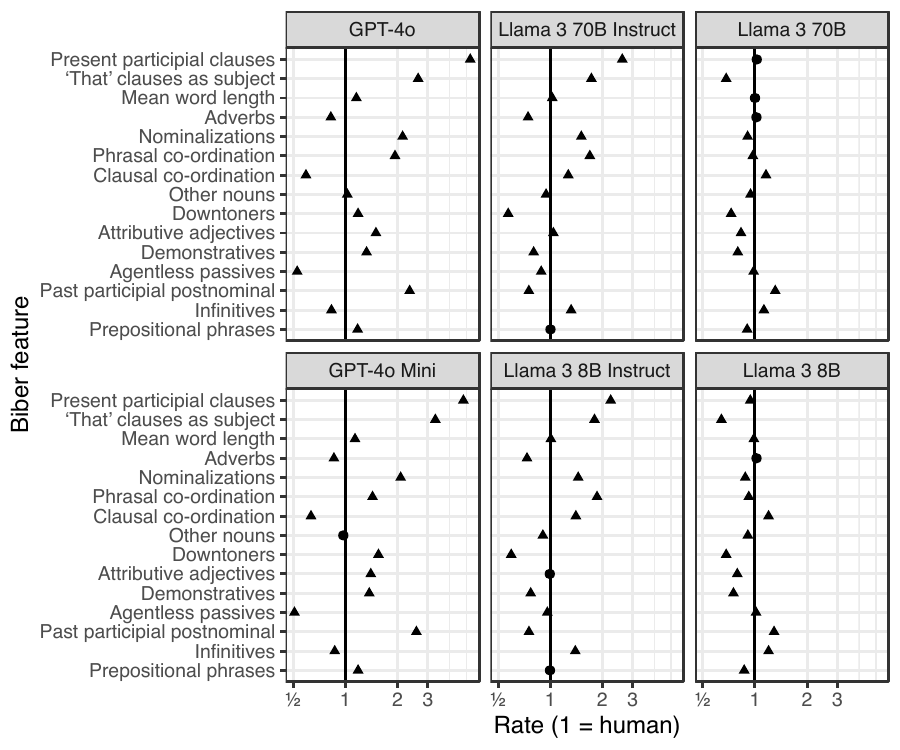}
    \caption{Rate of Biber feature use by different LLMs, relative to the human usage of each feature, for the top 15 most important features in the HAP-E corpus. Note the log scale. GPT-4o and GPT-4o Mini show the largest variation from human texts, while the base variants of Llama 3 most closely resemble human grammar and style. Larger models (top row) generally show the same stylistic differences as their smaller counterparts (bottom row), despite performing better on other benchmark tasks. Triangles indicate statistically significant differences from human usage. Figure S1 of the SI Appendix gives the distributions of paired differences.} 
    \label{fig:top-features}
\end{figure*}

Figure~\ref{fig:top-features} illustrates the large variation in rate of occurrence of the fifteen most important features (as identified by the random forest) in texts generated by LLMs, relative to the rate observed in the human text. All four instruction-tuned models have strong preferences for present participial clauses, `that' clauses as subjects, nominalization, and phrasal co-ordination, which are typical markers of more informationally dense, noun-heavy style of writing \cite{Aull:2020}. For example, GPT-4o uses present participial clauses  at 5.3 times the rate of humans (paired Cohen's $d = 1.38$), `that' clauses as subject 2.6 times as often ($d = 0.77$), nominalizations 2.1 times as often ($d = 1.23$), and phrasal coordination 1.9 times as often ($d = 0.81$). (Rates and effect sizes for all 66 features are provided in the SI Appendix, Tables S5--S6; Figure S1 illustrates paired differences.) There are also signs of local patterns that emerge with specific models: both GPT-4o models avoid clausal coordination, while all Llama 3 variants use it more frequently than humans; while both GPT-4o models use downtoners (such as \textit{barely} or \textit{nearly}) more frequently than humans, all Llama 3 variants avoid them.

One might expect larger models to better match human text than smaller models (e.g., Llama 70B versus Llama 8B, or GPT-4o versus GPT-4o Mini), but this does not appear to be the case in Figure~\ref{fig:top-features}. Also, instruction tuning appears to make the model output less human, not more: the Llama 3 base models use features at rates similar to human texts, while GPT-4o and Llama 3 instruction-tuned models have much wider variation from feature to feature. 

\begin{figure*}
    \centering
    \includegraphics[width=0.7\textwidth]{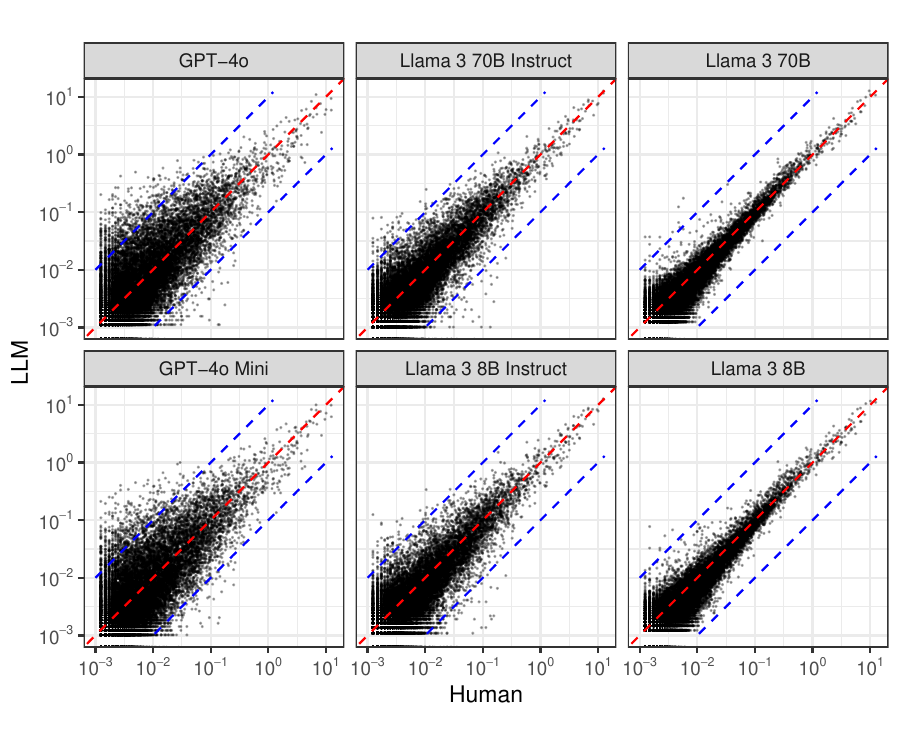}
    \caption{Rates of word use by different LLMs (per 1,000 words) compared to the human use of each word in chunk 2, in the HAP-E corpus (log scale). Includes all words used more than once per million words in chunk 2. Words are lemmatized to group together inflected forms. Words on the diagonal are used equally often in human and LLM texts. Dashed blue lines indicate the range between $10\times$ more and $10\times$ less than human use. Note that the instruction-tuned models show more variation from the diagonal, indicating more deviation in vocabulary use relative to humans.}
    \label{fig:vocab-compare}
\end{figure*}

Similar to past research \cite{Liang:2024}, we find that LLMs also favor specific vocabulary. Figure~\ref{fig:vocab-compare} shows the rate of usage for words used more than once per million words by humans, comparing the usage of each LLM to the usage by humans in Chunk 2 of HAP-E. Compared to the base Llama models, in the instruction-tuned Llama and GPT-4o models certain words are used at dramatically higher and lower rates. Table~\ref{tbl:overrepresented} highlights words overrepresented in LLM outputs: GPT-4o and 4o Mini use words like \textit{camaraderie}, \textit{palpable}, \textit{tapestry}, and \textit{intricate} at more than 100 times the rate of humans, such as in the GPT-4o output phrase ``The camaraderie was palpable.'' As a result, ``tapestry'' appeared in 23\% of GPT-4o outputs and ``amidst'' in 27\% (SI Appendix, Table S7). Instruction-tuned variants of Llama 3 also favor words like \textit{camaraderie} and \textit{palpable}, as well as \textit{unease} and \textit{reminder}, though at lower rates than GPT-4o and in a much smaller fraction of documents.\footnote{Some overuse may be artifacts of the generation process; for example, Llama 3 instruction-tuned variants overuse \textit{continuation} because their outputs sometimes begin with ``Here is the continuation of the text...'' Llama base models have a tendency to repeat themselves, so Llama 3 8B uses \textit{Deborah} at 52 times the rate of humans largely because of a single document repeating it 308 times.} Conversely, they use certain obscenities more than 100 times less often (SI Appendix Table S8).

\begin{table*}
\caption{Most overrepresented words in LLM-generated texts, relative to human usage rates}
\centering
    \begin{tabular}{lrlrlrlrlrlr}
\multicolumn{2}{c}{GPT-4o} & \multicolumn{2}{c}{GPT-4o Mini} & \multicolumn{2}{c}{Llama 3 70B Instruct} & \multicolumn{2}{c}{Llama 3 8B Instruct} & \multicolumn{2}{c}{Llama 3 70B} & \multicolumn{2}{c}{Llama 3 8B} \\ 
\cmidrule(lr){1-2} \cmidrule(lr){3-4} \cmidrule(lr){5-6} \cmidrule(lr){7-8} \cmidrule(lr){9-10} \cmidrule(lr){11-12}
Word & Rate & Word & Rate & Word & Rate & Word & Rate & Word & Rate & Word & Rate \\ 
\midrule\addlinespace[2.5pt]
camaraderie & 162 & camaraderie & 171 & unease & 63 & unease & 101 & bananas & 31 & deborah & 52 \\ 
tapestry & 155 & tapestry & 147 & palpable & 47 & continuation & 52 & paperback & 30 & rambo & 22 \\ 
intricate & 119 & palpable & 145 & continuation & 29 & palpable & 48 & bam & 26 & matty & 20 \\ 
underscore & 107 & grapple & 131 & shoutout & 28 & reminder & 33 & verona & 25 & goodnight & 18 \\ 
unspoken & 102 & intricate & 129 & intricate & 27 & pang & 29 & filth & 19 & ml & 15 \\ 
amidst & 100 & fleeting & 124 & pang & 25 & rut & 29 & rekall & 17 & merlin & 13 \\ 
palpable & 95 & ignite & 122 & camaraderie & 24 & waft & 28 & denis & 14 & worcester & 11 \\ 
solace & 95 & vibrant & 92 & policymaker & 24 & prioritize & 27 & darry & 12 & fay & 10 \\ 
fleeting & 84 & amidst & 90 & prioritize & 24 & grapple & 24 & ebook & 12 & missy & 10 \\ 
unravel & 83 & cacophony & 89 & reminder & 24 & camaraderie & 23 & janice & 12 & elisa & 10 \\ 
\bottomrule
\end{tabular}

\label{tbl:overrepresented}
\end{table*}

While many words listed in Table~\ref{tbl:overrepresented} may be occasionally expected in  belletristic works of fiction, their pervasiveness across LLM output in a diverse array of genres is notable. To those familiar with academic writing, newspapers, or television scripts, these words are largely unexpected, and to experts likely signal an overwritten, sentimental, or simply uneven text. The point here is not that humans refrain from using these words, but that humans refrain from using these words in certain genres. In this case, words that are unremarkable in fiction are highly conspicuous and unconventional when used other genres. As word choice appears most similar to humans for the base models, this suggests the word choice bias is introduced by the instruction tuning process, not simply by bias in the texts composing the training sets.

In the GPT-4o models in particular, many of these words connote some form of complex relation among objects (e.g., \textit{tapestry}, \textit{intricate}, \textit{camaraderie}, \textit{cacophony}, \textit{amidst}). Coupled with positive items such as \textit{vibrant} and \textit{solace}, these words together may signal a preference for grandiose, if hollow, summative sentences. 

\subsection*{Distinguishing individual LLMs}

When classifying between human-generated text and one specific LLM, rather than comparing all LLMs, our classifiers achieve much higher accuracy. Typical accuracies achieved by random forests were 93--98\% even when trained on HAP-E and tested on CAP or vice versa (SI Appendix Table S9).  Lasso-penalized logistic regression classifiers attained similar performance for all LLMs except for the Llama base models, which had accuracies around 75\% (SI Appendix Table S10). Since the lasso regressions only consider additive terms, this implies that interactions between the Biber features contain relevant signals for the Llama base models. 

For both methods, the lower classification accuracy for the Llama base models relative to the GPT-4o and instruction-tuned Llama models indicates that instruction tuning may lead to writing that is easier to distinguish from human writing.

\subsection*{Generalization across corpora}

When each pairwise random forest was used to classify arXiv preprints from the M4 corpus, accuracy dropped significantly. Random forests trained on instruction-tuned LLMs were able to classify M4's GPT-3.5 output with greater-than-chance accuracy, but models trained on the Llama base outputs attained only 50\% accuracy, equal to random guessing (SI Appendix Tables S11--S12). These results demonstrate that instruction-tuned LLMs do have features in common that permit their classification, but that generalizability across LLMs or to different registers of text is difficult.

\section*{Discussion}

This study identifies salient differences both between human and LLM-generated texts and among various models. The features that distinguish between humans and different LLMs include present participial clauses, `that' clauses as sentence subject, passive voice, and nominalizations, to name a few. These findings corroborate other research that points out the ways these features produce informationally dense prose \cite{Markey:2024}. Other research links these features to increased lexical diversity in generated text, as well as human judgments of linguistic mastery \cite{Herbold:2023iz}. Lastly, prior work found that ChatGPT-4 text evidences more nominalizations, and fewer human subjects and epistemic stance markers \cite{Jiang:2024sx}, findings we see reproduced in our list of distinguishing features.  

A second major finding of this research is the apparently central role of instruction tuning in creating these discrepancies between human and model general texts. While we do not have access to untuned versions of GPT, comparisons between Llama's base and tuned models emphasize the degree to which instruction tuning pushes models to produce text that reads \emph{unlike} a human. This suggests that differences in style are not simply due to the selection of texts for training the base models, but due to the instruction-tuning process. Similarly, differences between GPT-4o and the instruction-tuned Llama variants may be due to differences in instruction tuning, either through different human preferences in rating responses or differences in the tasks (such as summarization) used to tune the models. (As the instruction tuning processes are not publicly documented, it is not possible to determine the cause more precisely.) While instruction tuning has previously been shown to introduce cognitive biases \cite{Itzhak:2024}, to our knowledge, these changes in writing style are not discussed elsewhere in similar research. 

A third major finding of this work is significant success of Biber's tagset in modeling and classifying text. This success suggests that varied linguistic perspectives---which, perhaps, are not prioritized during the development and in-house assessments of LLMs---can reveal otherwise tacit information that distinguishes a text as machine-generated. With the linguistic perspective offered by pseudobibeR, we built a model that recognizes machine-generated text with relative ease. Our study reveals the clear value of linguistics expertise and functional conceptions of language in both LLM use and development. 

That said, our intention is not to propose another way to construct LLM detectors or to police the writing of students and learners. Instead, we maintain that this type of comparative analysis is useful for identifying differences between human- and machine-generated text, zeroing in on specific teachable moments in the revision of machine-generated text. 

As LLMs are increasingly put to work completing diverse writing tasks, these results suggest a notable misalignment between generated texts and the contexts in which we put them to use. This is another way of saying that LLMs do not vary their linguistic output in response to contextual factors in ways similar to humans. This misalignment affects experts and learners differently. For those proficient in a genre---think, a therapist collating notes on a patient or, say, a college graduate writing thank-you notes to friends and family---this misalignment is likely flagged and the output is appropriately revised. When the writer is proficient in the genre, previous experience guides a current sense of what a particular document should look like in order to be successful. For experts, then, LLMs appear a worthwhile productivity tool, suitable so long as they lend their expertise to further shaping the output. 

For learners, though, LLMs appear more problematic. Of course, using LLMs to learn more about a concept, or help generate ideas, is one thing. Using output in a text is a different matter, one that may affect a student's learning trajectories. In this case, students offload the important cognitive labor of shaping a text for a particular audience and purpose to the LLM. Never mind that LLMs do not appear to write like humans---when students offload writing work, they offload opportunities to learn how to write IMRaD articles, client reports, executive summaries, investment pitches, etc. When LLMs are used in the classroom as writing tools, instructors of all levels and disciplines need to help students see both the shortcomings of the generated text and avenues for improvement. LLMs are not bad, either technically or morally---it is only that instructors must help inculcate in students the critical perspective of the expert to know what's working and what isn't. 

In other contexts, however, overreliance on LLMs could produce output that might be awkward and inauthentic (e.g., in a creative genre), confusing (e.g., in instructional material), or unpersuasive (e.g., in argumentative texts). The current work thus suggests the importance of LLM practice---both in and out of the classroom---informed by human expertise via a continual dialogue of creation and revision, where LLM users are more aware and mindful of the effective uses as well as the limitations of various LLMs.

\matmethods{
}


\acknow{We thank members of the TeachStat Research Group for helpful discussions, the Dietrich College of Humanities and Social Sciences at Carnegie Mellon University for use of the Wright GPU cluster, and Aadi Menon for exploring suitable prompts.
}

\showacknow{} 

\bibsplit[4]

\bibliography{main}

\end{document}